\documentclass[pdflatex, sn-mathphys]{sn-jnl}

\usepackage{graphicx}%
\usepackage{multirow}%
\usepackage{amsmath,amssymb,amsfonts}%
\usepackage{amsthm}%
\usepackage{mathrsfs}%
\usepackage[title]{appendix}%
\usepackage{xcolor}%
\usepackage{textcomp}%
\usepackage{manyfoot}%
\usepackage{booktabs}%
\usepackage{algorithm}%
\usepackage{algorithmicx}%
\usepackage{algpseudocode}%
\usepackage{listings}%

\usepackage[acronym]{glossaries}
    \glsdisablehyper

\usepackage{tabularx}
\usepackage{makecell}

\newcolumntype{H}{>{\setbox0=\hbox\bgroup}c<{\egroup}@{}}

\theoremstyle{thmstyleone}%
\theoremstyle{thmstyletwo}%

\theoremstyle{thmstylethree}%

\begin{document}

\title[Article Title]{SynthEval: A Framework for Detailed Utility and Privacy Evaluation of Tabular Synthetic Data}


\author*[1]{\fnm{Anton D.} \sur{Lautrup}}\email{lautrup@imada.sdu.dk}

\author[1]{\fnm{Tobias} \sur{Hyrup}}\email{hyrup@imada.sdu.dk}

\author[1]{\fnm{Arthur} \sur{Zimek}}\email{zimek@imada.sdu.dk}

\author[1]{\fnm{Peter} \sur{Schneider-Kamp}}\email{petersk@imada.sdu.dk}

\affil[1]{\orgdiv{Department of Mathematics and Computer Science}, \orgname{Universitiy of Southern Denmark}, \orgaddress{\street{Campusvej 55}, \city{Odense}, \postcode{5230}, \country{Denmark}}}


\abstract{With the growing demand for synthetic data to address contemporary issues in machine learning, such as data scarcity, data fairness, and data privacy, having robust tools for assessing the utility and potential privacy risks of such data becomes crucial. SynthEval, a novel open-source evaluation framework distinguishes itself from existing tools by treating categorical and numerical attributes with equal care, without assuming any special kind of preprocessing steps. This~makes it applicable to virtually any synthetic dataset of tabular records. Our tool leverages statistical and machine learning techniques to comprehensively evaluate synthetic data fidelity and privacy-preserving integrity. SynthEval integrates a wide selection of metrics that can be used independently or in highly customisable benchmark configurations, and can easily be extended with additional metrics. In this paper, we describe SynthEval and illustrate its versatility with examples. The framework facilitates better benchmarking and more consistent comparisons of model capabilities.}




\maketitle

\section{Background and Motivation}
In recent years, the use of synthetic data has become increasingly popular for computer vision, machine learning, and data analysis in a wide range of domains. Synthetic data, which are generated using computer algorithms, offer several advantages over real-world data, such as reduced costs, augmented fairness, increased privacy, and the ability to simulate complex scenarios that may be difficult or impossible to capture in real life.

Personal level tabular records, especially, can be a treasure trove of valuable insights, holding personal information on patients, citizens, or customers and their various attributes, ripe for knowledge discovery tasks and new advanced predictive models to aid in decision-making and product development \citep{Davenport2019, Bhanot2021, Hernandez2022}. Conversely, tabular records also present a great challenge in terms of privacy - a Pandora's box that cannot and should not be easily unlocked. Data protection regulations may encumber scientific progress \citep{Ping2017, ElEmam2020} but are crucial in protecting the rights of individuals from malicious designs, blackmail, scams, or discrimination should sensitive data fall into the wrong hands \citep{Abouelmehdi2018}.

Synthetic data, in principle, allows tapping into the potential of tabular datasets and goes beyond traditional anonymisation techniques in protecting the privacy of individuals \citep{Nowok2016, Rankin2020}. Additionally, synthetic data can help alleviate data bias and make downstream models more robust, fair, and balanced \citep{Breugel2021}. While we have seen impressive results for text and image data, simulating tabular data still remains a challenge, and the quality and privacy of synthetic data can vary widely depending on the algorithms used to generate them, hyperparameter choices, and the evaluation framework. In particular, the lack of agreement on evaluation methods makes it arduous to compare existing models and thus difficult to objectively evaluate the performance of a new algorithm.

Several contemporary surveys \citep{Dankar2022, Figueira2022, Hernandez2022} call for standardised metrics and methods to evaluate and benchmark different aspects of synthetic data for these very reasons. Furthermore, privacy evaluation has been found to be lacking or absent from many studies. Differential Privacy (DP) is one method for ensuring privacy in the generation step \citep{Dwork2013} and does not strictly mandate post-generation evaluation. However, whilst DP may in principle provide a theoretical guarantee of privacy, research \citep{Yale2020, Lenz2021, ElEmam2022UM} suggests that the privacy level should still be evaluated, either to determine the appropriate parameters for DP or if only to be consistently comparable with methods that do not use DP. Multi-axis benchmarks are important to not only show the dimensions where a specific dataset is lacking but also to determine the best level of the privacy-utility trade-off. 

Consequently, in this paper, we propose a new evaluation framework, SynthEval\footnote{SynthEval is available from: \url{https://github.com/schneiderkamplab/syntheval}}, a Python library designed to make flexible and consistent evaluation of synthetic tabular data much more accessible. This will place an easily extendable library of evaluation metrics into the hands of researchers. To promote unbiased evaluation despite heterogeneous or mixed datatypes, in SynthEval, we propose taking special care to treat them in unison, using special distance functions or adapting metrics as required. We~expect that with SynthEval, highly detailed benchmarks will become more accessible, accelerating the process of evaluating synthetic data, and spurring the acceptance of new generative model frameworks.

In Section \ref{sec:relworks} below, we discuss the related works. Next, in Section \ref{sec:methods}, we introduce the components of the SynthEval framework. In Section \ref{sec:example} we apply the tool to an example before we in Section \ref{sec:concl} summarise, discuss, and conclude our work. 

\section{Related Work}\label{sec:relworks}
\begin{table}
\caption{\textbf{Different features of related works.} The table presents an overview of some readily available evaluation tools for evaluating tabular synthetic data. ``Ease of Use'' denotes how easily the tool can be integrated into an experimental setting. Configurability denotes if the tool allows bundling and customising metrics. Extendibility denotes if new custom metrics can easily be added, and Ranking denotes if the tool can help if multiple synthetic datasets are to be decided between. ``Mixed Type Strategy'' denotes the primary strategies the tool uses for taking care of numerical and categorical data. The number of metrics in SynthCity is approximate because a few of them are distinct implementations of the same metric (e.g. using different classifiers). For SynthEval, the ``18+12'' metrics signify that we have 18 different metric modules, but some of them produce supplementary results.}
\label{tab:contr}
\footnotesize
\setlength\extrarowheight{2pt}
\begin{tabular}{l | cccccl} 
    \toprule
    Framework & \# Metrics & Ease of Use & Config. & Extend. & Rank. & Mixed Type Strategy \\   
    \midrule
    SynthCity\tnote{\emph{a}} & $\sim24$ & $\star\star\star$ & $\star\star$ & \checkmark & - & Label enc. and binning \\
    SDmetrics\tnote{\emph{b}} & 17 & $\star$ & $\star\star$ & - & - & Separate \\
    TableEvaluator\tnote{\emph{c}} & 17 & $\star\star\star$ & $\star$ & - & (\checkmark) & Adapt and one-hot enc.\\
    DataSynthesizer\tnote{\emph{d}} & 2 & $\star\star$ & $\star$ & - & - & Binning\\
    \midrule
    SynthEval\tnote{\emph{e}} & 18+12 & $\star\star\star$ & $\star\star\star$ & \checkmark & \checkmark & Adapt (and separate) \\
     \bottomrule
\end{tabular}
\footnotetext{Granulation: Ease of Use; $\star\star\star$ few lines of code to get all necessary results, $\star\star$ minor work required e.g. dataset encoding or looping over metrics, $\star$ major preprocessing, coding, or additional files are needed. Config.; $\star\star\star$ metrics can be selected as desired AND metrics have accessible options, $\star\star$ either one of previous, $\star$ none of previous.}
\emph{a} \cite{Qian2023}, \emph{b} \cite{sdmetrics}, \emph{c} \cite{Brenninkmeijer2021}, \emph{d} \cite{Ping2017} , \emph{e} This work.
\end{table}

\begin{figure}
    \centering
    \includegraphics[width=0.7\textwidth]{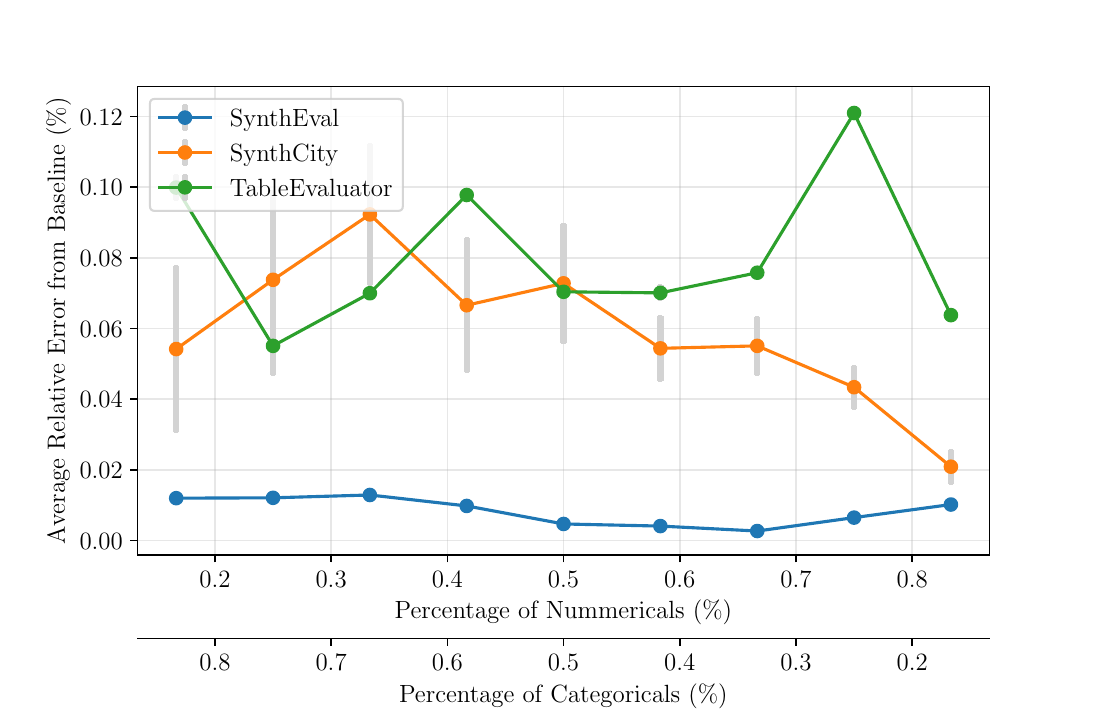}
    \caption{\textbf{Average Relative Error vs.~Attribute Type Proportion.} The figure shows how SynthEval, SynthCity, and TableEvaluator, each deviate from the baseline (all columns used) under repeated (10) subsampling of the categoricals and numericals in various mixtures using the same dataset (Egyptian Hepatitis C Dataset, see Table~\ref{tab:data}, and synthetic version generated using synthpop \citep{Nowok2016}). The metric used was the ``Similarity score'' in TableEvaluator and the sum of results from a selection of metrics in both SynthEval and SynthCity. Specifically; corr\_diff, mi\_diff, ks\_test, h\_dist, nnaa, eps\_risk, and dcr for SynthEval, and close\_values\_probability, chi\_squared\_test, feature\_corr, inv\_kl\_divergence, ks\_test, nearest\_syn\_neighbor\_distance, jensenshannon\_dist, max\_mean\_discrepancy and identifiability\_score for SynthCity (the last four were taken as one minus their value for the summation). SynthEval has only about $1\%$ deviation in all mixtures, whereas the TableEvaluator tool is only somewhat consistent in the intermediate mixtures at around $7$ to $8\%$ error, and fluctuates more in the extremes. SynthCity is less erratic but shows a steady decrease in error when fewer categorical columns are used.}
    \label{fig:stress}
\end{figure}






A few ecosystems for evaluating synthetic data already exist. Some that have usable Python implementations are listed in Table~\ref{tab:contr} above. Synthcity by \cite{Qian2023} hosts a variety of metrics in addition to an intuitive interface for accessing a selection of generative models. However, categoricals are treated as ordinal integers, making the evaluation slightly biased towards numerical values (see Figure~\ref{fig:stress}). Output values are mapped to the zero-one interval, which makes them clear to read but hides away some details. There is potential for confusion when metrics with the same units are mixed and some are to be ``maximised'' and others ``minimised''. SDmetrics\footnote{Also available as a module in their SDGym framework: \url{https://docs.sdv.dev/sdgym}} by \cite{sdmetrics} allows users to generate quality and diagnostic reports, as well as column-level figures for visual comparison. The selection of metrics is mostly limited to high-level dataset similarity and integrity with only a few recognised metrics appearing. The~framework requires a metadata file, which is non-trivial to create and compile, and was thus not included in the figure. Table Evaluator by \cite{Brenninkmeijer2021} is a library to evaluate how similar synthesised datasets are to real data through quantitative and visual evaluation. The evaluation is at large based on the performance of classifier models, together with a few other statistical methods. Metrics such as the correlation matrix are adapted to work for categorical data and one-hot encoding is used for nearest neighbour calculations. The tool produces a summary score that can be used for a quick ranking of datasets. The performance varies quite a bit for heterogeneous data. DataSynthesizer was developed by~\cite{Ping2017} and includes two metrics that check the quality of synthesised data visually using pairwise mutual information and marginal distributions. Numerical data are made discrete for this purpose by binning. 

There are other evaluation frameworks mentioned in the literature, for example, the work of \cite{Yale2019, ElEmam2022UM, Yan2022}, however, these are not supported by implementations that are readily applicable to new datasets.

SynthEval addresses several issues of its predecessors while integrating their strengths. Our evaluation framework is designed to be highly configurable and user-friendly, and it can easily be extended with custom metrics to suit specific domains or use cases. The current library of metrics is specially adapted to be as stable as possible on heterogeneous data, ensuring an unbiased and comparable evaluation (refer again to Figure \ref{fig:stress}). Furthermore, SynthEval avoids using a single summary statistic whilst allowing for easy dataset ranking and selection based on automatic multi-axis benchmarks with integrated ranking strategies.  

\section{Methods and Technical Solutions}\label{sec:methods}
\begin{figure}
    \centering
    \includegraphics[width=\textwidth]{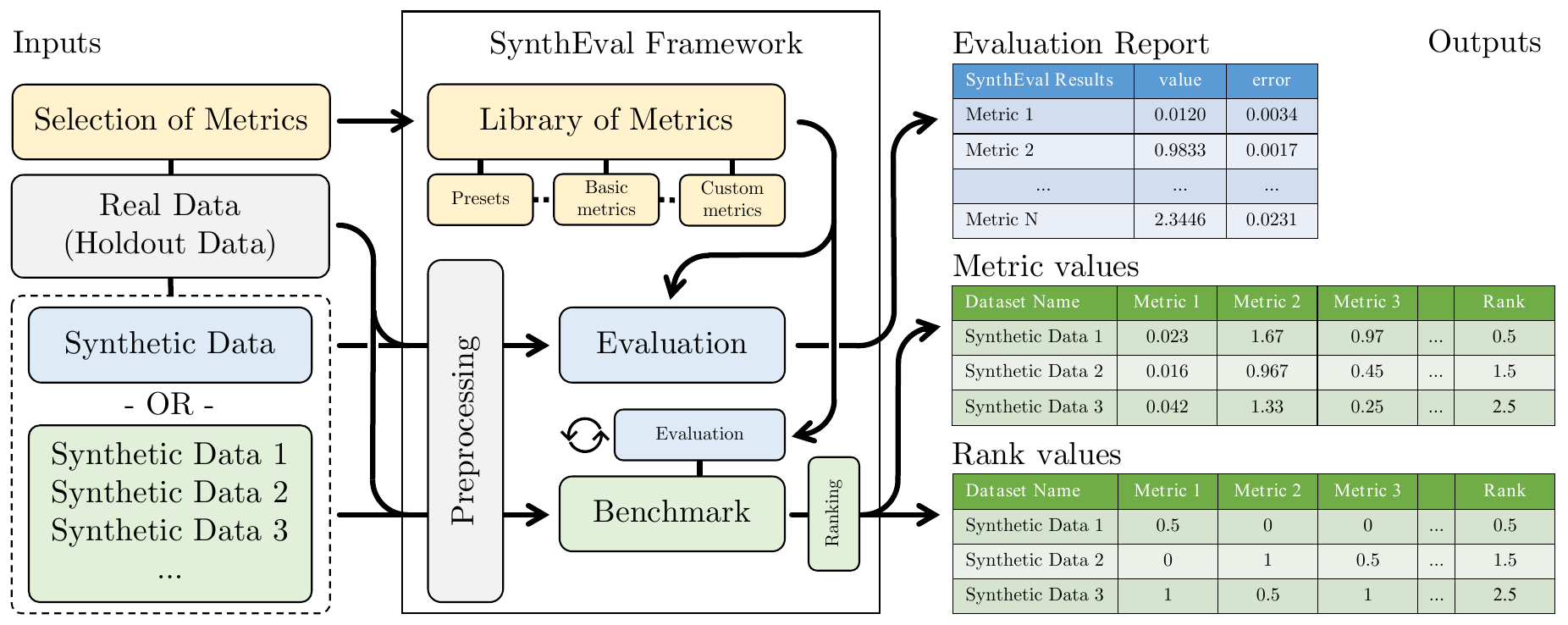}
    \caption{\textbf{Sketch of The SynthEval Framework.} The diagram shows the primary two workflows contained within SynthEval: single dataset evaluation and multi-dataset benchmarking. The standard evaluation module, allows the generation of detailed evaluation reports on a wide selection of metrics, using preset configuration and/or a manual selection of metrics (including custom metrics). The benchmark module enables the evaluation of multiple synthetic datasets simultaneously and returns the results in a joint table. Furthermore, the benchmark module ranks the results according to a specified ranking strategy, facilitating the identification of standout datasets. Finally, the individual metrics can also be accessed without entering the framework (not shown in the figure), requiring real data and synthetic data as inputs. In this configuration, the metrics can access the preprocessing utilities and take care of this step accordingly.}
    \label{fig:framework}
\end{figure}
In this section, we delve into the fundamental components and functionalities of SynthEval, an innovative framework tailored for flexible evaluation of tabular synthetic data. A rough sketch of the workflow can be seen in Figure~\ref{fig:framework}. We aim to provide a comprehensive overview of the core features underpinning SynthEval's architecture and capabilities in the following.

SynthEval is first and foremost a powerful shell for accessing a broad library of metrics and bundling them together into evaluation reports or benchmark configurations. This is achieved through two main components: the metrics object, and the SynthEval interface object. The former governs the structure of the metric modules and makes them available to the SynthEval workflow. The SynthEval interface object is the primary interactible object that hosts the evaluation and benchmark modules. The SynthEval utilities handle all required data preprocessing, including automatic determination of non-numerical values if they are left unspecified. In principle, evaluation and benchmarking require only two actions in code: the creation of the SynthEval object and the use of either method. We also make SynthEval available through the command-line interface. 

\subsection{Library of Metrics}
\begin{table}[t]
    \centering
    \setlength{\tabcolsep}{2pt}
    \caption{\textbf{Metrics Currently Available in the SynthEval Tool.} The key column shows metrics' names in custom evaluation configurations, to the right are the available options for the metric implementation. The ``types'' column denotes what datatypes the metric considers, those marked by~* were adapted as part of this work to function on all datatypes, and those marked by~$^\dagger$ use the nearest neighbour utility described in the text. The range indicates what the ideal value of the metric is best (leftmost) to worst (rightmost). Finally, $\#A$ refers to the number of attributes in the data. For more details on these metrics, see the supplementary material.}
    \label{tab:mets}
    \begin{tabular}{>{\ttfamily}r l c c l}
        \toprule
        key & metric name & types & range & options \\
        \midrule
        \multicolumn{4}{l}{\textsc{Utility metrics}}\\
        \midrule
        dwm         & Average Dimension-wise Means Difference    & num. & [0, 1) &  \\
        pca         & Principal Component Analysis Visualisation& num. & -    & \\ 
        cio         & Average Confidence Interval (CI) Overlap  & num. & (1, 0] & confidence=95 \\
                    & -- Number of Non-Overlapping CIs          &     & [0, \#A] & \\
                    & -- Fraction of Non-Overlapping CIs        &     & [0, 1] & \\
        corr\_diff  & Correlation Matrix Difference             & all* & [0, $\infty$) & mixed\_corr=True \\
        mi\_diff    & Mutual Information Matrix Difference      & all  & [0, $\infty$) & \\
        ks\_test    & Average Kolmogorov-Smirnov (KS) Distance  & all  & [0, 1) & sig\_lvl=0.05,   \\
                    & -- Average P-Value                        & all* & (1, 0) & -- n\_perms=1000 \\
                    & -- Number of Significant Tests            &   & [0, \#A] & \\
                    & -- Fraction of Significant Tests          &   & [0, 1] & \\
        h\_dist     & Average Empirical Hellinger Distance      & all* & [0, 1] & \\
        p\_mse      & Propensity Mean Squared Error (pMSE)      & all  & [0, 0.25) & k\_folds=5,  \\
                    & -- Average pMSE Classifier Accuracy       &      & [0, 1) & -- max\_iter=5000\\
        nnaa        & Nearest Neighbour Adversarial Acc. (NNAA) & all$^\dagger$  & [0, 1) & n\_resample=30 \\
        auroc\_diff & Binary prediction AUROC difference        & all  & [0, 1) & model='log\_reg' \\
        cls\_acc     & Average Difference in F1 Score (train)   & all  & [0, 1) & F1\_type='micro', \\
                    & -- Average Difference in F1 Score (test)  & all  & [0, 1) & -- k\_folds=5\\
        \midrule
        \multicolumn{4}{l}{\textsc{Privacy metrics}}\\
        \midrule
        nndr        & Nearest Neighbour Distance Ratio (NNDR)   & all$^\dagger$ & (1, 0) & \\
                    & -- Privacy Loss (diff. in NNDR)           &      & [0, 1) & \\
        nnaa        & Privacy Loss (diff. in NNAA)              &      & [0, 1) & \\
        dcr         & Median Distance to Closest Record (DCR)   & all$^\dagger$ & ($\infty$, 0) & \\
        hit\_rate   & Hitting Rate                              & all*  & [0, 1] & thres\_percent=1/30 \\
        eps\_risk   & Epsilon Identifiability Risk              & all$^\dagger$ & [0, 1] & \\
        mia\_risk   & Membership Inference Attack (MIA) macro F1& all   & [0, 1] & num\_eval\_iter=5 \\
                    & -- Average Precision                      &      & [0, 1] & \\
                    & -- Average Recall (MIA Risk)              &      & [0, 1] & \\
        att\_discl  & Attribute Disclosure Risk (ADR) macro F1  & all  & [0, 1] & \\
                    & -- Average Precision                      &      & [0, 1] & \\
                    & -- Average Recall                         &      & [0, 1] & \\
        \bottomrule
    \end{tabular}
\end{table}
Currently, SynthEval contains 18 separate metric modules that can be combined into multifaceted benchmarks according to the user's specifications. The metrics currently available are shown in Table~\ref{tab:mets}, listed with the additional 12 outputs. The metrics were carefully selected based on trends in recent research literature and surveys \citep{Dankar2022, ElEmam2022UM, Hernandez2022, Yan2022, Murtaza2023, Lautrup}, in the hope that these metrics, when made easily available, can establish themselves further together. All metrics are additionally configurable and can be accessed individually, circumventing the SynthEval interface without any loss of control.

\subsubsection{Heterogeneous Data} 
An integral design philosophy in building the SynthEval framework is that it should be as applicable as possible, working on both numerical and categorical data types in any mixture without any significant bias. This required some major refinements in the integration of certain metrics, and there are now only three that treat solely the numerical attributes (two of which are primarily for visual verification).

Most notably was the creation of the Nearest Neighbour (NN) utility, which allows metrics that require finding the NNs to use the same settings. As the default option, we elect to use Gower's similarity/distance measure \citep{Gower1971} rather than Euclidean distance for all NN computations. Gower's distance averages normalised similarity scores/distance measures across attributes using type-appropriate metrics. This approach offers a key advantage: there is no need for one-hot encoding of categorical variables (unlike Euclidean distance), which avoids overwhelming the memory with numerous binary variables. Furthermore, Gower's distance relies on normalised metrics, eliminating concerns about dominant signals from large variables, and eliminating the need for assumptions regarding data normalisation. Should the user prefer the more transparent but potentially misleading option of using the Euclidean distance, this can be enabled as a global setting in the framework.

Several more specific refinements are present throughout the metrics library, such as empirical approximation of p-values and construction of mixed-type correlation matrix analogues. The reader is encouraged to consult Appendix \ref{app:mets} for details. 

\subsubsection{Pre-configurations}
In order to be as flexible as possible, SynthEval offers several ways to access the included metrics. Metrics can either be hand-picked from the library and/or selected in bulk by activating one of the three currently implemented preset configurations. If~any non-standard configuration is used, a new config file will be saved in JSON format for reproducibility (if a file path is used as a preset, SynthEval will attempt to load the file). The available presets currently comprise \texttt{full\_eval}, \texttt{fast\_eval}, and \texttt{priv\_eval} tailored to different use cases: Full evaluation, which activates all metrics in SynthEval. Fast evaluation, which is designed to have only the more scalable/fast metrics and is therefore suitable for scenarios such as checkpoint evaluation during model training or for larger datasets. Privacy evaluation activates only the privacy metrics, which are useful if domain-specific utility metrics have already been run externally or if only the privacy of the data is important. 

\subsubsection{Extensibility}
The SynthEval framework is highly modular and allows for easy local integration of new and domain-specific metrics for immediate use. We hope this will enable new metrics to appear alongside more recognised evaluation methods, ultimately increasing the trust and transparency of synthetic data evaluation. The metrics object and the one-file-one-metric philosophy provide a simple way to interface with SynthEval. It~only requires following a convenient template\footnote{\url{https://github.com/schneiderkamplab/syntheval/blob/main/src/syntheval/metrics/metric_template.py}} and placing the new metric file in the appropriate directory. If done correctly, the metric can now be used locally. We~hope to encourage a growing collaborative platform for integrating metrics from a multitude of domains based on user input and contributions.  

\subsection{Benchmark Module} 
The benchmark module is another of SynthEval's convenient features that allows several synthetic versions of the same dataset to be evaluated together. The results are pooled, ranked internally, and exported. This allows the user to comprehensively evaluate a wide selection of datasets on a broad selection of metrics with little effort. Internal ranking deals with the pitfall of information overwhelm in multi-axis comparison and trade-off, enabling researchers to quickly identify datasets that perform well and on what metrics, but also without hiding away all the raw data. Additionally, this approach is more compelling than having a single summary score for each evaluation report such as an aggregate measure, since rank-derived scoring does not certify performance outside of the benchmark setting but only contrasts results within~\citep{Yan2022}. 

\subsubsection{Ranking Strategies} 
SynthEval offers three ways to conduct ranking, differing mostly in how the intermediate results are treated: ``linear'', ``normal'', and ``quantile''. The ``linear'' ranking strategy, which is the default, imposes min-max normalisation on the results of each metric, indicating the proximity of intermediate results to the best/worst result. This~scheme is more appropriate for datasets that differ more from each other in terms of overall quality, perhaps produced through different generative approaches rather than from the same model with different hyperparameters. The ``normal'' strategy assigns a score of $1$ to the best metric result and $0$ to the worst. Intermediate results are all scored $0.5$. This scheme works to separate the overall best and worst datasets, when the results are closely grouped, and where we may not be able to say much objectively about the intermediate results subject to noise. Finally, the ``quantile'' strategy sorts the results into four quantiles and assigns them scores of 0, 1, 2, and 3. This~schema can assign multiple winners and losers in each metric, which is perhaps more fair than assigning a single best and worst result. Accordingly, this schema is only appropriate when many datasets are being compared at once. 

After ranking the results of each metric, the dataset profiles are added together to produce the overall ranking (the result is unweighted). Additionally, a separate dataframe displaying all the metric-level scores is provided to aid in interpreting the raw values (see Figure~\ref{fig:framework}).

\section{Application of SynthEval}\label{sec:example}
There are many potential applications of the SynthEval Framework such as in-depth model comparisons and dataset benchmarks. Because the former is out of scope for this publication, here, as a prototypical example, we demonstrate how our framework can be employed to rigorously assess and compare synthetic datasets created by different generative processes. Ultimately, this allows for guiding the creation and selection of a high-quality synthetic dataset that corresponds to the user's specifications. 

\subsection{An Example of Dataset Benchmarking}
\begin{table}[]
    \centering
    \caption{\textbf{Egyptian Hepatitis C Dataset Details}. The dataset has no missing values and includes $29$ variables across $1{,}385$ observations. The variables marked as discretised were made categorical based on the expert recommendations in the dataset supplement file. We chose not to discretise all of the numerical attributes but only those with extreme values (attributes with ranges that reached thousands and more).}
    \label{tab:data}
    \setlength\extrarowheight{2pt}
    \begin{tabular}{r l}
    \toprule
    type & attributes \\
    \midrule
    numerical & \makecell[tl]{Age, BMI(Body Mass Index), HGB(Hemoglobin), AST1(1 week), ALT1(1 week), \\ ALT4(4 weeks), ALT12(12 weeks), ALT24(24 weeks), ALT36(36 weeks), \\ ALT48(48 weeks), Baseline Histological Grading.} \\
    binary & \makecell[tl]{Gender, Fever, Nausea/Vomiting, Headache, Diarrhea, Fatigue, Bone ache,\\ Jaundice, Epigastria pain.} \\
    categorical & Baseline Histological. \\
    discretised & \makecell[tl]{WBC(White Blood Cells), RBC(Red Blood Cells), Plat(Platelet), RNA Base, \\ RNA 4, RNA 12, RNA EOT, RNA EF(Elongation Factor).} \\
    \bottomrule
    \end{tabular}
\end{table}
A commonly acknowledged potential application of synthetic data is in unlocking personal-level microdata, such as health records, for analysis or replication by third-party analysts outside of the dataset's parent organisation~\citep{Rankin2020, Bhanot2021}. For such a data analysis to be successfully outsourced, the synthetic data needs to provide a valid statistical analysis while protecting privacy.

For the current demonstration, we select the ``Hepatitis C Virus (HCV) for Egyptian patients'' dataset from UCI\footnote{\url{https://archive.ics.uci.edu/dataset/503}} \citep{Kamal2019}, which, although publicly available, is a good toy model for the kind of datasets from life sciences that could be of interest to third-party researchers. Additionally, it has a suitable size for this experiment and has a good mix of numerical and categorical variables (after using the supplied discretisation criteria on the variables with extreme ranges -- see Table~\ref{tab:data}). For the sake of the example, let us assume that a research group at a hospital is interested in the ``Baseline Histological'' class, which, based on the other parameters, signifies what stage of disease a patient is in (four levels). The group wishes to employ a new exciting AI solution, available at a third-party organisation, to learn if they can improve decision-making in a clinical setting. To safely share the data with the data analysts, the hospital's in-house data scientists prepare synthetic versions of the dataset. For the train-test split, we choose $2/3$ of the data for training and keep a hold out of $1/3$ for evaluation. 

\subsubsection{Preparation} 
Before generating synthetic data through any one approach, it is crucial to establish a baseline for the experiment. With tabular synthetic data, a baseline can mean many things, such as noisy copies of the real data, using the holdout data, synthetic minority oversampling technique (SMOTE) \citep{Chawla2002} samples, or creating questionable synthetic data by sampling features independently. 

To set a good example, we perform several of these baselines: SMOTE, independent random sampling of the marginals, and a ``deletion and imputation'' baseline where we remove half of the entries in each column at random and replace them with the mean value (or most frequent observation for categoricals). 

\subsubsection{Generating Synthetic Data}
For generating the synthetic data, we make use of four readily available synthetic data generators. We use two generative adversarial network models, ADS-GAN~\citep{Yoon2020} and CTGAN~\citep{Xu2019} implemented through synthcity~\citep{Qian2023}, we use DataSynthesizer~\citep{Ping2017} for a Bayesian network implementation, and sequential classification and regression tree models from Synthpop~\citep{Nowok2016} in~R.

Generative adversarial networks (GANs) are deep learning models where two sub-model neural networks compete against each other; the generator attempts to create synthetic samples that will pass as real using only random noise and the discriminator endeavours to distinguish between real and synthetic samples \citep{Goodfellow2014}. These models are challenging to train but are also one of the more promising deep-learning approaches for tabular synthetic data. ADS-GAN adds a privacy term to its training cost function, and CTGAN changes numericals encoding into Gaussian mixture models, which makes modelling multimodal distributions more natural. 

Bayesian networks (BN) are graph models based on directed acyclic graphs, where each variable is a node and the directed edges represent variable relationships. Sampling from a Bayesian network is probabilistic and uses the learned conditional probability tables \citep{Sun2015}. DataSynthesizer is a quite popular implementation of BNs that also supports differential privacy (disabled by default). Classification and regression tree (CART) models similarly synthesise the variables in some order and use simple classification or regression tree models to create each of the variables~\citep{Reiter2005, Drechsler2011}. The concept is simple but effective, and the Synthpop library is a prominent implementation of this approach. 

Due to the stochastic nature and instability of some of the generation algorithms, we trained three models for each method and only recorded the best result (using SynthEval benchmark, with the ``normal'' ranking strategy). In all cases, the synthetic datasets have $1{,}000$ samples as the target size to be of comparable size to the training data. 

\subsubsection{Model Tuning and Selection} 
After running the models without any adjustment, we noticed that all the synthetic datasets produced unsatisfactory results for privacy. In the next step, hyperparameter optimisation, we wish to remedy this problem and attempt to make each model perform better on privacy while not disregarding utility. Namely, we wish to bring down the $\epsilon$-identifiability risk seen in Table \ref{tab:results}, closer to the recommended level\footnote{Few data authorities have concrete guidelines, yet the \cite{EuropeanMedicinesAgency2018} and \cite{HealthCanada2019} have recommended aiming for an identification risk of less than 9\%.}, and hopefully also close the gap to the baseline privacy rank. In some sense, the first round of synthetic datasets were a second baseline to see what the models could do on their own without spending time on hyperparameter tuning. We carry out a non-exhaustive (10 rounds) random search of the parameter space and record the result that improves the privacy level the most without entirely disregarding utility. 

\subsubsection{Evaluation and Comparison}
\begin{sidewaystable}
\setlength\extrarowheight{2pt}
    \setlength{\tabcolsep}{2pt}
    \centering
    \caption{\textbf{SynthEval Benchmark Results.} The table shows the result of running SynthEval, on the eleven synthetic datasets used in this study, with a custom configuration of metrics. The rows are ordered by total rank, within the three subcategories of datasets (baseline, initial results, privacy boosted). The numbers are rounded to two significant figures, or to the first digit of the error where appropriate*}\label{tab:results}
    \begin{tabular*}{\textheight}{@{\extracolsep\fill}l ccccccc cccc ccc}
        \toprule%
        & \multicolumn{7}{@{}l@{}}{\textsc{Utility metrics}} & \multicolumn{4}{@{}l@{}}{\textsc{Privacy metrics}} & \multicolumn{3}{@{}l@{}}{\textsc{Rank}}\\\cmidrule{2-8}\cmidrule{9-12}\cmidrule{13-15}%
        
        & Corr. & MI & KS dist.* & KS frac. & pMSE* & \makecell[c]{F1 diff.\\ train}* & \makecell[c]{F1 diff.\\test}* & mDCR & $\varepsilon$ risk & MIR* & ADR* & tot. & U & P \\ \midrule
        \multicolumn{4}{l}{\textsc{Baseline results}}\\ \midrule
        random sample & 2.5 & 2.4 & \textbf{0.017}(2) & \textbf{0}/29 & 0.0036(3) & 0.023(9) & 0.018(7) & 1.1 & 0.28 & 1.0(0) & 0.41(6) & 7.0 & 5.4 & 1.6 \\
        SMOTE & \textbf{0.41} & 3.0 & 0.07(1) & 13/29 & 0.049(0) & \textbf{0.011}(10) & 0.026(8) & 0.82 & 0.53 & 0.71(0) & 0.46(7) & 4.9 & 4.6 & 0.30 \\
        del. \& impute & 2.1 & 2.8 & 0.26(3) & 27/29 & 0.065(2) & 0.19(0) & 0.18(3) & 1.2 & \textbf{0.060} & \textbf{0.0}(0) & 0.36(7) & 4.2 & 0.59 & 3.6 \\ \midrule
        \multicolumn{4}{l}{\textsc{Initial model results}}\\ \midrule
        BN (DS) & \textbf{0.41} & \textbf{0.17} & 0.020(3) & 2/29 & 0.0004(2) & 0.030(10) &	\textbf{0.017}(10) & 0.97 & 0.51 & 1.0(0) & 0.45(7) & 7.3 & \textbf{6.8} & 0.49 \\
        CART (sp) & 0.44 & 0.27 & 0.018(2) & \textbf{0}/29 &	\textbf{0.0003}(2) & 0.025(9) & 0.030(12) & 0.97 & 0.54 & 1.0(0) & 0.44(7) & 7.2 & \textbf{6.8} & 0.42 \\
        CTGAN & 0.81 & 1.9 & 0.052(8) & 11/29 & 0.020(0) & 0.030(11) & 0.025(11) & 1.0 & 0.51 & 0.85(1) & 0.44(7) & 6.1 & 5.3 & 0.78 \\
        ADSGAN & 0.83 & 1.9 & 0.047(6) & 14/29 & 0.015(0) & 0.032(10) & 0.040(7) & 0.97 & 0.53 & 0.83(1) & 0.44(7) & 5.8 & 5.1 & 0.68 \\ \midrule
        \multicolumn{4}{l}{\textsc{Privacy boosted model results}}\\ \midrule
        CART (sp) & 0.56 & 0.20 & 0.020(3) & \textbf{0}/29 & 0.0010(3) & 0.028(10) & 0.030(16) & 0.96 & 0.54 & 1.0(0) & 0.44(7) &	7.2 & 6.7 & 0.46 \\
        BN (DS) & 2.3 & 2.2 & 0.14(3) & 12/29 & 0.063(3) & \textbf{0.011}(11) & 0.03(2) & \textbf{1.6} & 0.062 & 0.19(1) & \textbf{0.35}(5) & \textbf{7.4} & 3.6 & \textbf{3.8} \\
        CTGAN & 2.5 & 2.8 & 0.063(11) & 16/29 & 0.043(2) & 0.013(11) & 0.046(9) & 1.1 & 0.29 & 0.92(0) & 0.41(6) & 5.5 & 3.8 & 1.7 \\
        ADSGAN & 3.4 & 2.8 & 0.098(13) & 21/29 & 0.071(3) & 0.025(11) & 0.036(17) & 1.3 & 0.27 & 0.70(2) & 0.42(7) & 4.8 & 2.8 & 2.0 \\ \botrule
    \end{tabular*}
    \footnotetext{Shorthands:  MI -- Mutual Information, KS -- Kolmogorov Smirnov, pMSE -- propensity Mean Squared Error, mDCR -- median Distance to Closest Record, MIR -- Membership Inference Risk, ADR -- Attribute Disclosure Risk, U -- Utility, P -- Privacy.}
    \footnotetext[*]{Metrics that are averages of several results, e.g., across each attribute, or from cross-validation. The standard error is marked with parenthesis.}
\end{sidewaystable}
We use SynthEval's benchmark module with a custom evaluation configuration and the linear ranking strategy\footnote{The ranking is performed over all 11 synthetic datasets, to keep everything on the same scale. The few lines of code for performing this full analysis are found in the guide on the SynthEval GitHub repository \url{https://github.com/schneiderkamplab/syntheval/blob/main/guides/syntheval_benchmark.ipynb}.}. The results are arranged into Table~\ref{tab:results}, which closely resembles the output from SynthEval. The baselines seem to offer lots of competition for the generative models, especially the ``random sample'' seems a strong baseline with one of the highest total ranks, and a combination of privacy and utility scores that are not surpassed simultaneously throughout the benchmark.

Looking at the results in Table~\ref{tab:results}, it is apparent that ensuring high utility does not imply good privacy. The datasets with best utility (unoptimised BN and CART models) also have some of the worst rankings in terms of privacy, with unacceptable identification risks. All models decreased in performance when privacy was up-prioritised, only the dataset created using the BN DataSynthesizer algorithm with differential privacy parameter $\varepsilon=0.18$ ended up with a somewhat acceptable balance. Most of the privacy metrics are close to the lowest in the entire table and $\epsilon$-identifiability risk is within the acceptable margin. Other metrics, such as the membership inference risk and attribute disclosure, however, are not, illustrating the pitfall of blindly relying on differential privacy for ensuring privacy. In other words, this dataset could not be relied upon to protect sensitive information, and more rounds of optimisation should be performed in practice.  

Now, if we, for the sake of the example, accept this non-ideal privacy level, the other interesting question becomes: How much is the utility of the data impacted? The~utility rank of the optimised BN dataset is the third lowest in the entire table. Is~this actually a problem that would make it unusable? In order to make this judgement, we have to look at the constituent metrics. 

At first glance, we note that the practical utility measured with both of the F1 differences is actually close to the best values in the entire table, which is an interesting result, as many possible applications of the synthetic dataset may involve prediction models of the ``Baseline Histological'' class: the fact that we will not be much worse off using the synthetic data that also best (besides the extreme baseline) protect privacy is encouraging. 

\begin{figure}
    \centering
    \includegraphics[width=0.49\textwidth]{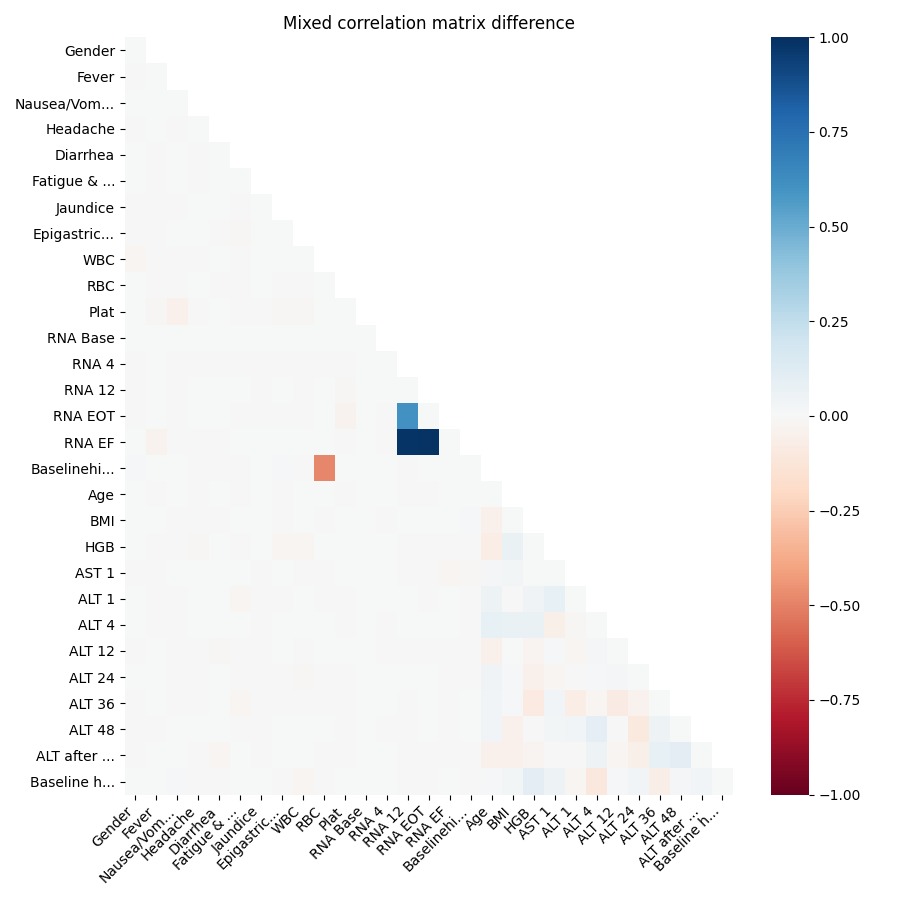}
    \includegraphics[width=0.49\textwidth]{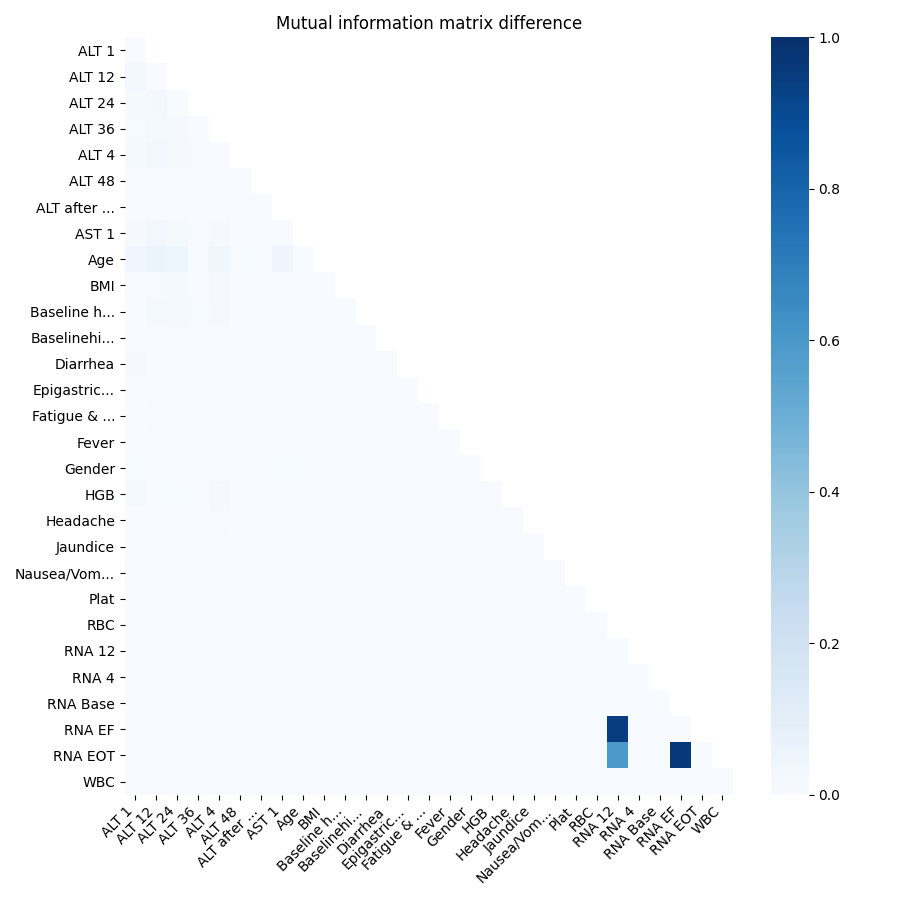}
    \caption{\textbf{Mixed Correlation and Mutual Information Matrix Difference for the Optimised BN}. Both plots are produced by SynthEval. Left: The figure shows the mixed correlation difference map of the real and synthetic data. It is evident that although the BN model had one of the worst correlation difference coefficients in Table \ref{tab:results}, it is only because some few variable interplays have been misrepresented. Right: The figure shows the mutual information difference map of the real and synthetic data. Some of the same misrepresentations of variable relationships seen on the left are found with this approach.}
    \label{fig:corr_mi}
\end{figure}

Next, looking at the correlation coefficient difference and the mutual information map; they are not close to zero, but also not too large (considering the number of attributes). Scrutinising the full difference maps (see Figure \ref{fig:corr_mi}), reveals that the problem seems to be that a few of the categorical attributes have had their relationships slightly mismodelled. Namely, RNA EOT, RNA EF and RNA 12. Whether this is a detrimental problem can be considered, and if the variables are not, after all, important for the downstream analysis, it may be reasonable to remove them.

\begin{figure}
    \centering
    \includegraphics[width=\textwidth]{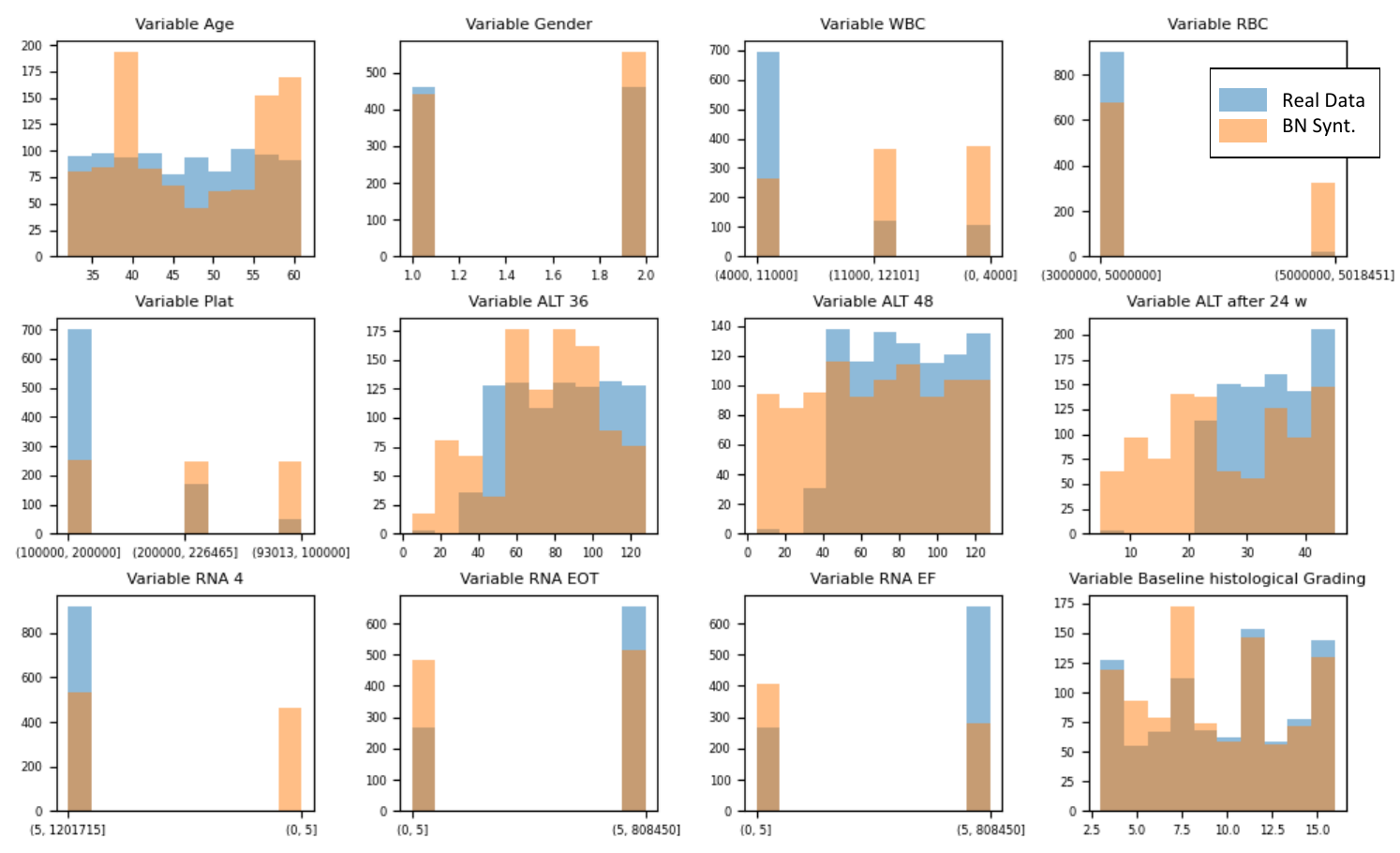}
    \caption{\textbf{Kolmogorov-Smirnov test, significantly dissimilar variables of the optimised BN dataset.} This figure shows how the values in the 12 variables that the KS test identified as significantly dissimilar are distributed. As is evident, with these hyperparameters the optimised BN generative model tends to balance the categorical attributes and to fill the gaps between outliers and the main population for numericals.}
    \label{fig:ks_hist}
\end{figure}

Another measure that reveals concrete but potentially fixable problems is the fraction of significant Kolmogorov-Smirnov tests. From the statistical and empirical p-values (p-values are estimated with resampling for categoricals), we can identify columns that are not populated as in the real dataset. Investigating these variables in Figure~\ref{fig:ks_hist}, we notice that the optimised BN model with its current hyperparameter settings tends to balance the categoricals that are imbalanced in the training data, and populates the empty space for numerical variables from outliers to the main distribution. Perhaps it is a good thing to have more samples in some of these columns, especially when the correlations are not unreasonable. However, some of these attributes were also the culprits from before, which may give us more reason to remove them. 

Identifying possible ways forward in this scenario could require either some domain-informed decision or going back to hyperparameter tuning. Either way, we leave this application example here.

\subsection{Sketch of Model Benchmark}
\begin{figure}[t]
    \centering
    \includegraphics[width=\textwidth]{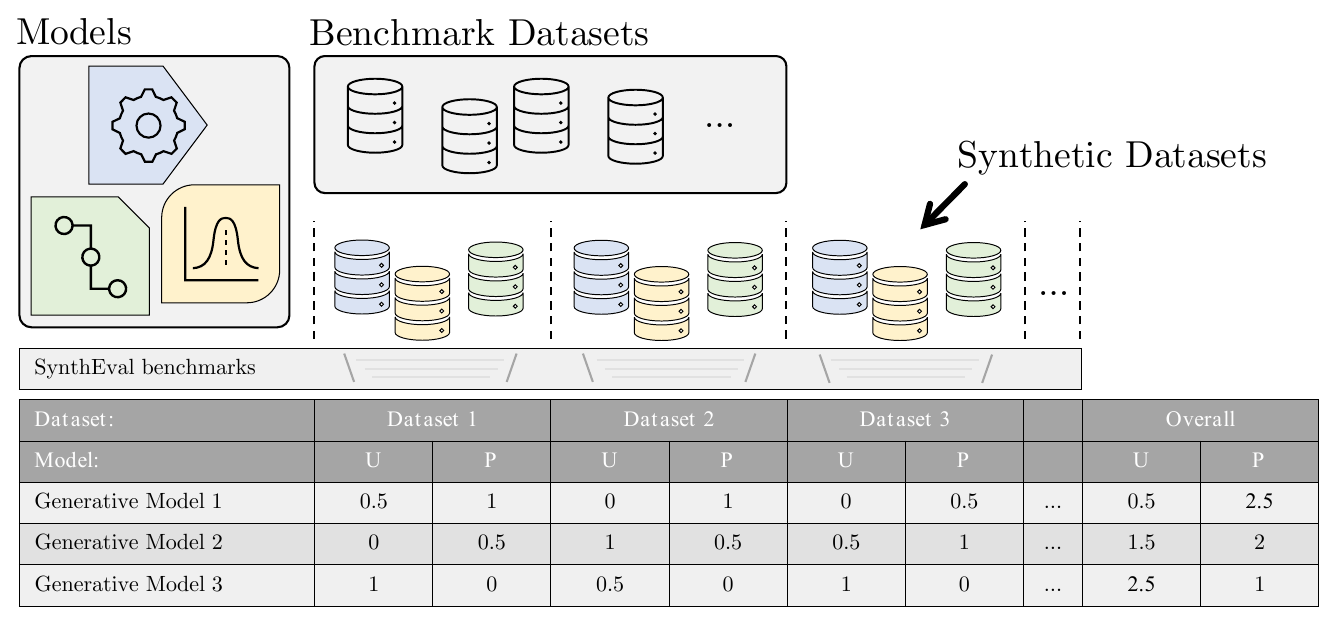}
    \caption{\textbf{Outline of Model Benchmark Study Design.} A selection of generative models is applied to a large collection of benchmark datasets. The resulting synthetic datasets are ranked internally on utility and privacy, and the results across all the different benchmark datasets are aggregated, to select the model that best solves the problem specification. It may be worthwhile to also look into subclusters of the benchmark datasets, as some models may perform better/worse on special niches, e.g., large/small datasets or datasets with lots of binary variables etc.}
    \label{fig:model}
\end{figure}
In this section, we outline one possible approach for conducting a larger model selection/evaluation benchmark. There are several other excellent approaches for conducting such an experiment: \cite{Yan2022} test their models on two datasets, closely analyse several metrics, and produce a rank-derived score for each of their models. \cite{Dankar2022} try their models on a wide palette of datasets and look for the model with the lowest classification accuracy loss across all datasets and four classifiers. 

For conducting a rigorous model benchmark study, we recommend putting together a wide selection of datasets of different categories (including large/small datasets, few/many columns or combinations thereof) and generating synthetic versions. The~SynthEval benchmark module provides a detailed breakdown of the models' proficiencies in adapting each of the real datasets. Comparing the models' ranking on each dataset or across the board and within each category allows the researcher to determine whether and where each model performs well and whether a specific type of model performs better on a certain subspace of datasets (see Figure~\ref{fig:model}). It is important to consider that just because a model places well regarding utility or privacy metrics it does not necessarily imply that it is useful or private. The results should be understood in relative rather than absolute terms, i.e. we should only conclude that it performs better in comparison to the others. 

\section{Concluding Remarks}\label{sec:concl}
With the increased focus on synthetic data in health and other application areas, it is increasingly important to have reliable and thorough evaluation methodologies to verify the utility and privacy of synthetic data. Frameworks such as SynthEval lay the foundation for detailed benchmarks of the capabilities of generative models through the datasets they create. A great challenge from the perspective of tabular data is how to handle the changing proportions of categorical and numerical data, as well as to provide consistent results. Previous evaluation frameworks have avoided this issue in different ways, either by restricting the types of data they accept or by limiting the available metrics. SynthEval, however, takes a different approach and attempts to capture the intricacies of real data by accommodating heterogeneity through the use of similarity functions as opposed to classical distances, empirical approximation of p-values, and constructions of mixed correlation matrix analogues.

We are inspired by the hope that the SynthEval framework will bring a much-needed solution to the evaluation scene and have a substantial impact on the research community. SynthEval integrates a broad library of metrics that can be assembled into benchmarks tailored to the user's preferences. Presets offer standardised benchmarks easily available with the activation of a keyword, while custom configurations can be easily built using the provided components. Custom metrics can be created with relative ease, and be included in benchmarks without any changes to the existing source code.

In terms of limitations, the metrics available in SynthEval each account for dataset heterogeneity in their own way. There is only so much that can be achieved with preprocessing, and future metric integrations will have to be mindful of conforming to dataset heterogeneity as a major challenge in synthetic data. It is the intention that SynthEval will be maintained: We hope to keep increasing the performance of the various algorithms and the framework already implemented and plan to add further metrics requested or contributed by the community. 

\backmatter

\bmhead{Supplementary information}
The SynthEval framework is available on \href{https://github.com/schneiderkamplab/syntheval}{GitHub} with update log and user guide or directly through \href{https://pypi.org/project/syntheval/}{PyPI}. 

\bmhead{Disclosure of potential conflicts of interest} 
Arthur Zimek is serving as action editor for DAMI.

\bmhead{Acknowledgments}
This study was funded by Innovation Fund Denmark in the project ``PREPARE: Personalized Risk Estimation and Prevention of Cardiovascular Disease''.


\bibliography{bibliography}


\begin{appendices}

\section{Metrics Theory and Implementation}\label{app:mets}
In this supplement, we briefly outline the technical details of the metrics that are supplied alongside the current version of SynthEval (see Table \ref{tab:mets}) and declare any unconventional choices in their implementations.

The nearest neighbours module mentioned in a few places below, refers to a piece of the SynthEval framework that allows control over what distance metric is used in distance calculations. Currently, the options are Euclidian distance and Gower's similarity\footnote{Using \url{https://github.com/wwwjk366/gower}}, but more may be available in future versions. 

\subsection{Utility Metrics}

\begin{figure}[!htb]
    \centering
    \includegraphics[width=\linewidth]{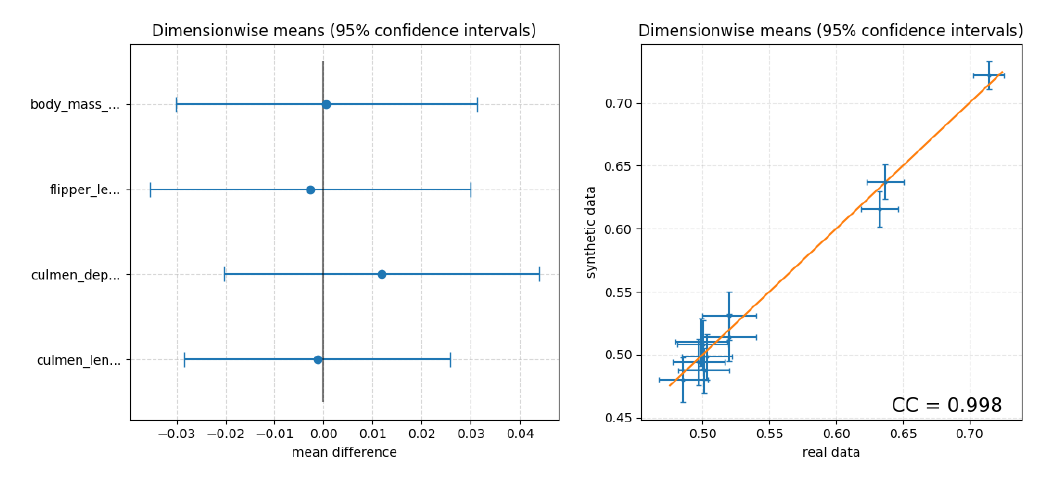}
    \caption{\textbf{Dimension-wise Means Figures.} The figure shows the two different plots the \texttt{dwm} metric provides. Left: Plot generated when there are fewer than ten numerical attributes. If Zero is contained within the confidence intervals, the attribute is modelled acceptably. Right: For ten or more numerical attributes the diagram plots the mean of the real attribute plotted against the synthetic attribute. If~the diagonal is contained in the confidence intervals the attribute is modelled acceptably.}
    \label{fig:dwm}
\end{figure}

\paragraph{Average Dimensionwise Means (DWM) Difference.} Measures the average difference in (min-max normalised) means of numerical attributes, between the real and synthetic data. Smaller is better. The implementation also provides a figure (different figures for less than ten and ten and more numerical attributes), see Figure \ref{fig:dwm} for an example.  

\begin{figure}[!htb]
    \centering
    \includegraphics[width=\linewidth]{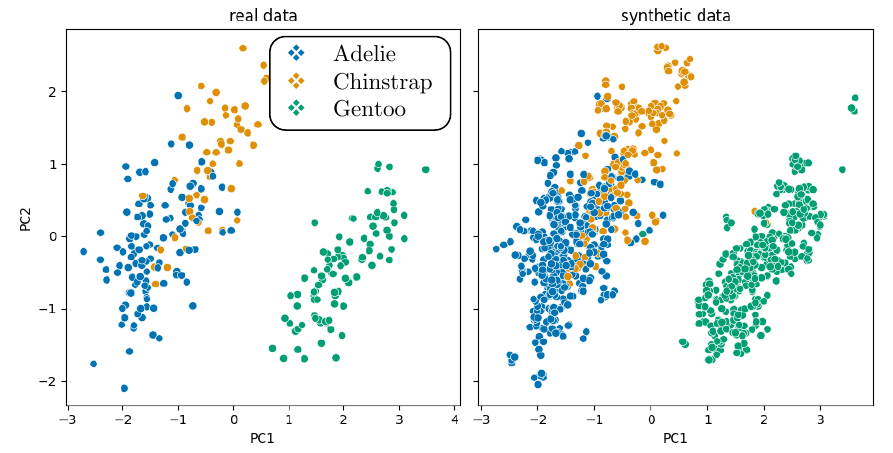}
    \caption{\textbf{Principal Component Analysis Figures.} Shows the real and synthetic data projected into the same principal component space.}
    \label{fig:pca}
\end{figure}

\paragraph{Principal Component Analysis Visualisation.} A visualisation of the numerical attributes mapped projected along the two first principal components. If the distributions are not remarkedly dissimilar, it is good. See~Figure \ref{fig:pca} for an example. 

\paragraph{Average Confidence Interval Overlap (CIO).} Metric that measures the average percentage overlap of the 95\% confidence interval estimates of numerical attributes. Higher is better, see \cite{Karr2006} for details. We also provide the number and fraction of non-overlaps, here lower is better.

\begin{figure}[!htb]
    \centering
    \includegraphics[width=0.9\textwidth]{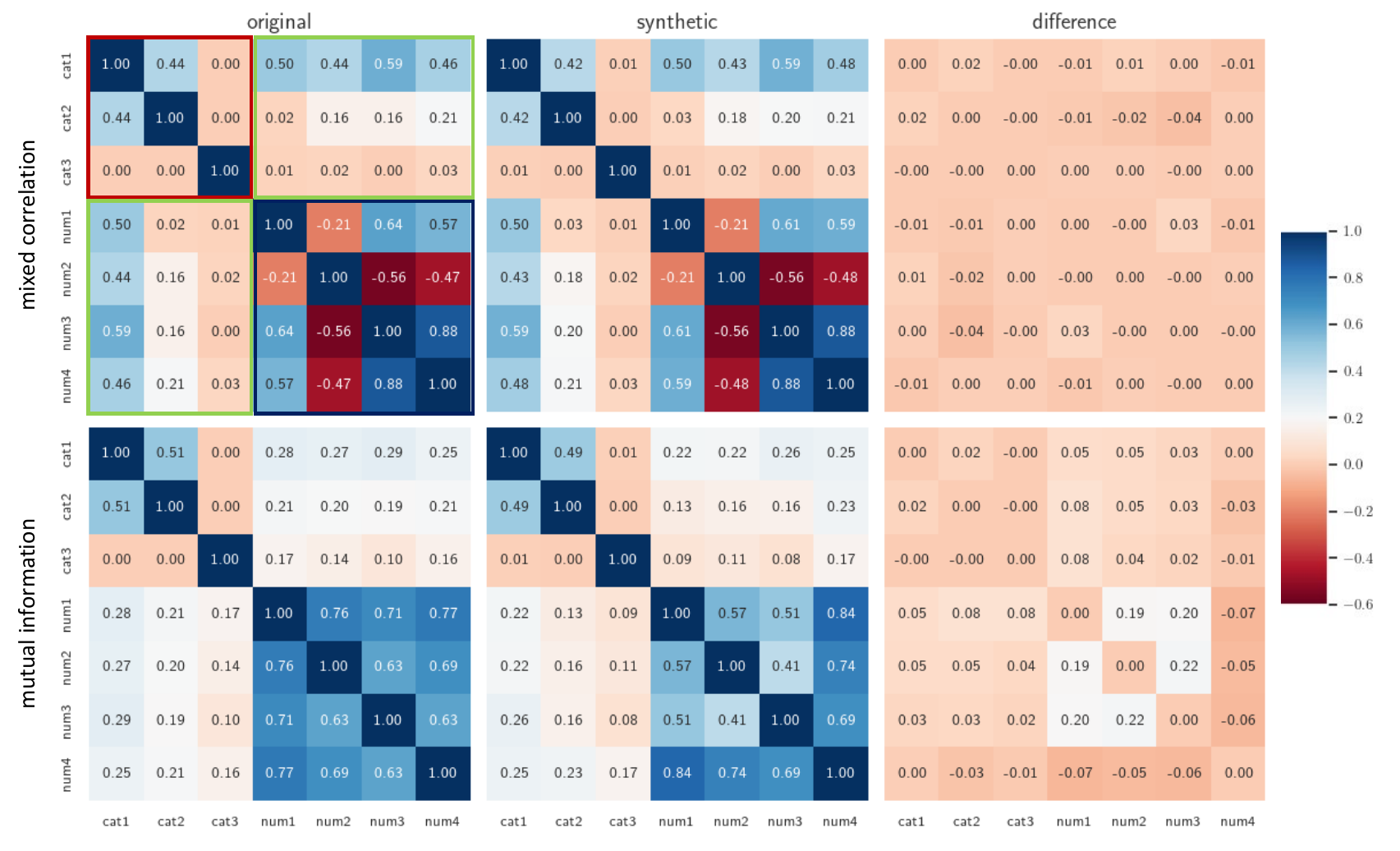}
    \caption{\textbf{Mixed Correlation and Mutual Information Matrices.} Top row is the ``correlation'' matrices constructed using Pearson's correlation coefficient (blue outline), Cramer's V (red outline) and correlation ratio (green outline). The bottom row is the mutual information matrix. The leftmost column shows how the matrices look for the training data, the centre column is a synthetic version (made using DataSynthesizer \cite{Ping2017}), and the rightmost column holds the difference map. The Frobenius norm of the difference matrices here is 0.1037 and 0.5595, for correlation and mutual information, respectively.}
    \label{fig:mat_diff}
\end{figure}

\paragraph{Correlation Matrix Difference.} As exemplified in the main paper, this metric provides a difference heatmap of the real and synthetic empirical correlation matrices. We calculate the numerical-numerical correlations using Pearson's correlation coefficient, the categorical-categorical correlations using Cramer's V, and the categorical-numerical correlations using correlation ratio~$\eta$ as they suggest in \cite{Zhu2022}, annotated example is seen in Figure \ref{fig:mat_diff}. The single number result is calculated by taking the Frobenius norm of the difference matrix, but the full matrices can be obtained through options. Lower values are better.

\paragraph{Mutual Information Matrix Difference.} Metric that behaves very similarly to correlation above, except for not needing special treatment for categorical data. The SynthEval implementation borrows from \cite{Ping2017} that is based on the \texttt{sklearn.metrics} function \texttt{normalized\_mutual\_info\_score}. An example is seen in Figure \ref{fig:mat_diff} along the bottom. Again the single number is calculated by taking the Frobenius norm of the difference matrix. Lower is better.

\begin{figure}[!htb]
    \centering
    \includegraphics[width=0.8\textwidth]{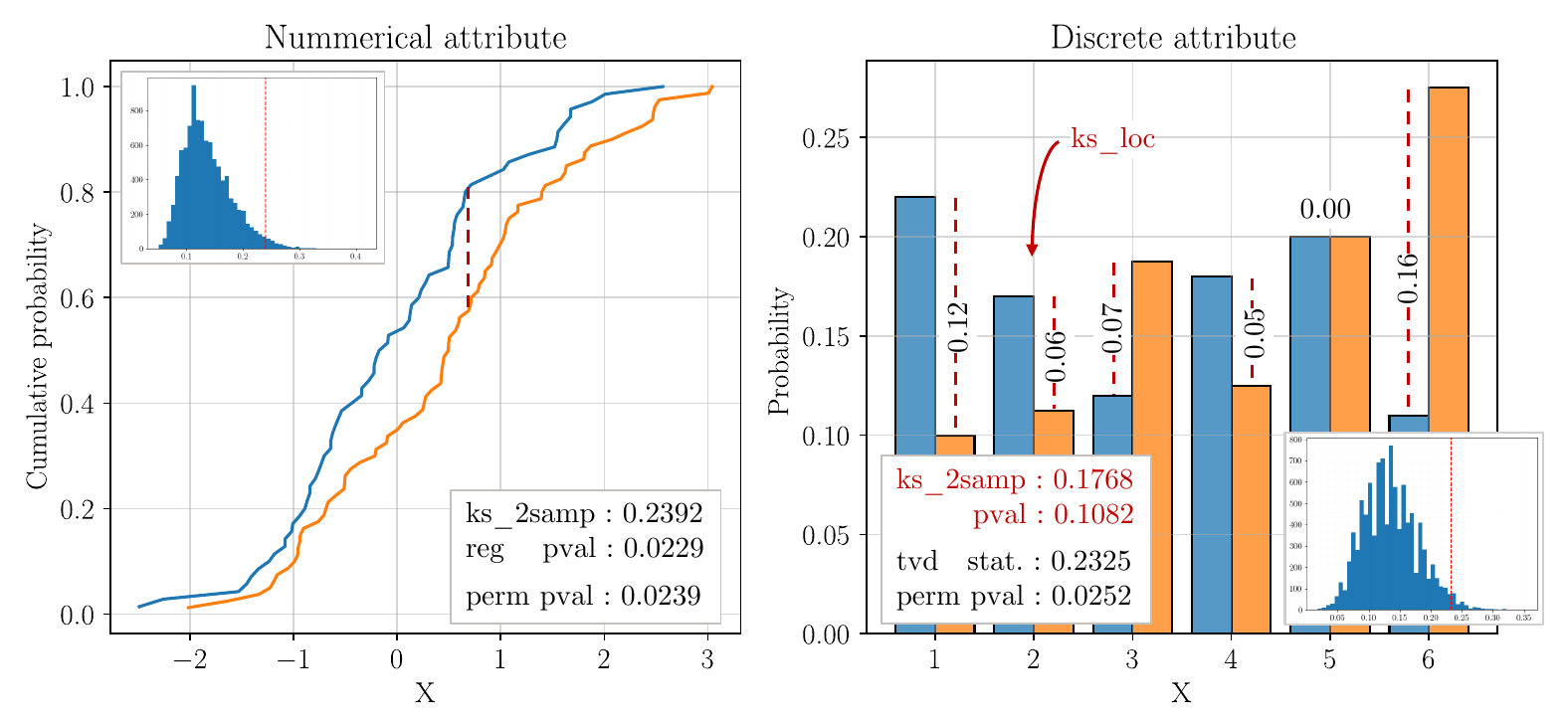}
    \caption{\textbf{Calculation of Kolmogorov-Smirnov statistic} Left: KS-statistic for samples from a continuous distribution. The p-value obtained from \texttt{ks\_2samp} is close to agreeing with that from the permutation test. Right: Histogram of a discrete attribute. Using \texttt{ks\_2samp} would not work very well, since it is dependent on the category order, which can be arbitrary. In this example, the test provides a maximum separation of 0.18 on the second category, due to the cumulative signal of the first two population differences. Using Total Variational Distance instead will consistently result in the same value (sum of disparities divided by two) even if we change the order of the categories. Under a permutation test, with $10{,}000$ random reassignments, the TVD reveals that population differences are unlikely to occur by chance. The numerical and discrete examples have the same number of samples.}
    \label{fig:ks_ecdf}
\end{figure}

\paragraph{Kolmogorov-Smirnov (KS) / Total Variation Distance (TVD) Test.} This metric tests whether marginal distributions are significantly dissimilar, in terms of location and shape. It consists of the Kolmogorov-Smirnov distance and Total Variation Distance for numerical and categorical attributes, respectively. The former is a hypothesis test commonly used in the evaluation of synthetic data \cite{Dankar2022}. The statistic is interpretable as the maximum separation of the two empirical cumulative distribution functions (eCDF, see Figure~\ref{fig:ks_ecdf}). The implementation of the KS test used in SynthEval is the \texttt{scipy.stats} implementation \texttt{ks\_2samp}. A discrete version of the KS test can be performed but will provide spurious test statistics due to the non-ordinal nature of categorical variables. Instead, Total Variational Distance is implemented for categorical attributes, since for these, the order of levels does not affect the outcome. The TVD summarises the disparity between populations in each category and thus captures similarity in terms of shape and location irrespective of order \cite{Villani2009}: 
\begin{equation}
    \mathrm{TVD}(P,Q) = \frac{1}{2}\sum_x|P(x)-Q(x)|,
\end{equation}
in terms of probability mass functions $P$ and $Q$. We obtain an empirical p-value by using permutation\footnote{The number of permutations can be adjusted. By default, the test performs 1000 resamplings, corresponding to a false positive rate of about 1 in 500.} \citep{Hesterberg2009}. We return the average combined test statistic (low values are preferred), the average KS and TVD statistics, the average combined p-value (to help with interpretation), the number and fraction of significant tests (better if low), and a list of the significant columns. 
 
\paragraph{Average Empirical Hellinger Distance.} Another widely used measure of univariate distributional similarity (distance)\citep{Dankar2022,ElEmam2022UM,Lenatti2023}. In SynthEval we discretisise the numerical attributes to use the discrete version of Hellinger distance. We determine appropriate binning by Scott's reference rule\footnote{Scott's normal reference rule determines the bin width using $3.49{s}/\sqrt[3]{n}$ where $s$ is the sample standard derivation and $n$ is the number of observations.} \citep{Scott1979}. The attribute-wise distances are averaged. Lower is better. 

\paragraph{Propensity Mean Squared Error (pMSE).} Measures how well a classifier model (in SynthEval we use standard \texttt{LogisticRegression} from \texttt{scikit-learn}) can distinguish real and synthetic data from each other~\citep{Woo2009, Raab2017, Snoke2018, Hornby2021}. It is calculated as an average based on 5-fold cross-validation (by default) and uses the formula described in~\cite{Woo2009}. A low value is better, worst-case is 0.25. We also output the classifier accuracy since it is sometimes used as an additional metric.

\paragraph{Nearest Neighbour Adversarial Accuracy (NNAA).} Metric that theoretically attempts to quantify how well a competent classifier should be able to differentiate the synthetic and real data~\cite{Yale2020}. This metric uses the NN-module in SynthEval to calculate nearest neighbours. A~lower value is better, but too low raises some overfitting concerns that may be assessed by evaluating NNAA on holdout data (see privacy losses below). The metric is a little unstable towards datasets of different sizes, so we sample equal-sized batches if the synthetic data outnumber the real samples by more than a factor of two (repeat 30 times by default). 

\paragraph{Binary Prediction AUROC Difference.} 
This metric only works when the supplied attribute to be predicted has two levels, otherwise, it disables itself. The metric has the option to enable subsampling from the datasets in order to make a more robust curve with confidence intervals (for the visualisation). The metric checks the difference in the area under the receiver operator curve for a model trained on real data vs. one trained on fake and evaluated on real hold-out data. The metric also draws a plot. 

\paragraph{Average Difference in F1 Score.} This metric shows how well four different classifiers do on classifying new real samples if they are trained on real or synthetic data, and measures the difference in accuracy (F1~micro, by default). We use the scikit-learn's decision tree, AdaBoost, random forest, and logistic regression classifiers, for both internal dataset 5-fold-cross-validation, but also allow the user to supply holdout data for testing that were not used in training the generative model. A~lower value is better.

\subsection{Privacy Metrics}

\paragraph{Nearest Neighbour Distance Ratio (NNDR).} Metric for calculating the point-wise nearest neighbour distance (using the SynthEval NN-module) of synthetic to real samples and compares it to the next-nearest neighbour distance \citep{Ooko2021}. Higher is better, but its privacy loss is perhaps more meaningful. 

\paragraph{Privacy Losses.} Measure the difference in a metric as measured on the training data and on a holdout test set. The values indicate if the synthetic data match the training data more than they match the new, never-before-seen data and, therefore, warn of overfitting and loss of privacy~\citep{Yale2020, Ooko2021}. Better if lower, negative counts as zero\footnote{Neither of the papers describing privacy losses comment on this. Because it is unreasonable to expect that the synthetic data are more like the holdout data than the real data, we assume no loss of privacy if privacy losses are below zero. It may be worth checking the holdout data for information leakage.}.

\paragraph{Median Distance to Closest Record (DCR).} Metric with many different interpretations in the literature \citep{Fan2020, Zhao2021, Ooko2021}, here we check the ratio of the median distance from the synthetic samples to the nearest real (using the NN-module), to the median distance of real samples to their own nearest neighbours,
\begin{equation}
    \mathrm{med.DCR}=\frac{\mathrm{med}\{\min_{r\in R} d(s,r) | s \in S\}}{\mathrm{med}\{\min_{r'\in R\setminus r}d(r,r')| r\in R\}}.
\end{equation}
The measure ranges from 0 and is unbounded from above, values below 1 indicate synthetic data being closer to real records than other reals, while a larger value indicates that we are further away from overfitting. 

\paragraph{Hitting Rate.} Rather naive membership inference risk metric \citep{Yan2020}, that checks if we have any exact matches from the real data in the synthetic data. To have a little wiggle room for the numerical values, we define exact matches with a 1/30 threshold of the variable range (as in \cite{Fan2020}, corresponding to about 3\%). This threshold can be changed. We record the fraction of real samples that were ``copied'' at least one time. The lower the value the better. As mentioned in the main paper, it is customary to aim for an ``identification'' risk below 9\%. 

\paragraph{Epsilon Identifiability Risk.} Ratio of real samples that have a synthetic point closer than the next real sample. This uses the NN-module in SynthEval but weighs the samples by inverse column entropy to provide greater attention to rare data points \citep{Yoon2020}. A synthetic dataset is considered to be $\epsilon$-identifiable if the proportion of real samples $R$ that are too similar to a neighbouring synthetic record $S$ is less than $\epsilon$. A~lower value is better, again below 9\% is a good first target.

\paragraph{Membership Inference Attack (MIA)}
This metric simulates a membership inference attack. Membership disclosure occurs when an adversary has access to data from the population the synthesis was made from and is able to infer whether some of the known records have used for training~\citep{ELEmam2022MD}. In this implementation, we assume that the adversary has no knowledge about the generation process, but access to the resulting synthetic data and real data from the same population. The MIA checks the performance of a model (random forest classifier) trained to distinguish the synthetic data from a population of some external data (in this case requiring holdout data). The model's ability to identify whether test samples of mixed real and holdout data fall into either category tells us about identification risks in the data. In this scenario, the model's recall is analogous to the membership inference risk.

\paragraph{Attribute Disclosure Risk (ADR)}
Attribute disclosure risk denotes the risk that an adversary with a partial dataset holding fewer attributes than the synthetic data can use the synthetic data to infer information about missing variables. The implementation in SynthEval assumes a worst-case scenario, where an adversary has knowledge of real records with the same attributes present in the synthetic data with the exception of a single attribute. A~predictive model is trained on the synthetic data to predict the target $\mathcal{T}$ using the remaining attributes as predictors $\mathcal{P} = \{x | x \in \mathcal{S} \setminus \mathcal{T} \}$ where $\mathcal{S}$ denotes the entire synthetic dataset. To provide the disclosure risk for all attributes, we cycle through the attributes and train an adversarial model based on the synthetic predictors $\mathcal{P}$, to infer the missing value, $\mathcal{T}$. The average accuracy of the adversary denotes the total attribute disclosure risk.

To provide a combined attribute disclosure risk for both categorical and numerical attributes, separate approaches must be considered. For categorical targets, a random forest classifier is trained, and the adversary's prediction accuracy is recorded on the ordinally encoded categories. Computing the adversarial accuracy for numerical targets is slightly different. A random forest regressor is trained using the synthetic predictors and the min-max normalised target. If the absolute difference between real, $y$ and predicted value, $\hat{y}$, is less than or equal to a predefined threshold, $\tau$, the predictions is recorded as a hit. Formally, we defined the numerical accuracy for a single attribute as:

$$\text{Acc}_{\text{num}} = \frac{1}{n} \sum_{i=0}^n \mathbb{I}(|y_i - \hat{y}_i| \leq \tau )$$
where n is the number of records in the synthetic data, $\mathcal{S}$, and $\mathbb{I}$ is the indicator function. The default threshold in SynthEval is set to $\tau = \frac{1}{30}$ similar to that of the hitting rate~\cite{Fan2020}.




\end{appendices}

\end{document}